\documentclass[letterpaper, 10 pt, conference]{ieeeconf}  

\IEEEoverridecommandlockouts               

\usepackage{cite}
\usepackage{amsmath,amssymb,amsfonts}
\usepackage{algorithmic}
\usepackage{graphicx}
\usepackage{textcomp}
\usepackage{xcolor}
\usepackage{caption}
\usepackage{amssymb}
\usepackage{amsmath}
\usepackage{amsbsy}
\usepackage{pgfgantt}
\usepackage{bbm}
\usepackage{graphicx}
\usepackage{subfigure}
\usepackage{wrapfig}
\usepackage{xspace}
\usepackage{xcolor}
\usepackage{dsfont}

\definecolor{wine}{RGB}{204, 0, 102}
\definecolor{ocean}{RGB}{13, 121, 202}
\definecolor{light_ocean}{RGB}{18, 178, 235}
\definecolor{dark_ocean}{RGB}{10, 89, 148}
\definecolor{grey}{RGB}{170, 170, 170}
\definecolor{light-grey}{RGB}{220, 220, 220}
\definecolor{dark_gray}{rgb}{0.2, 0.2, 0.2} 
\definecolor{grape}{RGB}{112,48,160}
\definecolor{aqua}{RGB}{52,172,139}
\definecolor{dark_aqua}{RGB}{35,115,93}
\definecolor{dark_orange}{RGB}{216,92,0}
\definecolor{vibrant_orange}{RGB}{255, 102, 0}
\definecolor{vibrant_blue}{RGB}{14, 120, 255}
\definecolor{vibrant_pink}{RGB}{255, 0, 104}
\definecolor{dark_red}{RGB}{122, 0, 0}
\definecolor{dark_green}{RGB}{0, 92, 34}

\usepackage[
  bookmarks=false,
  pdfpagelabels=false,
  hyperfootnotes=false,
  hyperindex=false,
  colorlinks,urlcolor= orange,
  citecolor=orange,
  linkcolor=orange,
]{hyperref}

\usepackage{tikz}
\usepackage{booktabs}
\usetikzlibrary{shapes}
\usetikzlibrary{calc} 
\usepackage[compact]{titlesec}
\titlespacing{\paragraph}{%
  0pt}{
  0.2\baselineskip}{
  .5em}
\usepackage{pgfplots}
\usepgfplotslibrary{fillbetween}	

\newcommand{\para}[1]{\medskip\noindent\textbf{#1. }}

\newcommand{\parasi}[1]{\smallskip\noindent\textit{\textbf{#1. }}}

\newcommand{\ours}{\textcolor{orange}{\textbf{SALT}}\xspace}
\newcommand{\rl}{\textcolor{grape}{\textbf{RewardSum}}\xspace}
\newcommand{\ensemble}{\textcolor{gray}{\textbf{Ensemble}}\xspace}

\newcommand{\Euclid}{\textcolor{dark_gray}{\textbf{Euclidean}}\xspace}
\newcommand{\SIRL}{\textcolor{wine}{\textbf{SIRL}}\xspace}
\newcommand{\GPT}{\textcolor{light_ocean}{\textbf{LLM}}\xspace}
\newcommand{\BERT}{\textcolor{dark_aqua}{\textbf{CosineSim}}\xspace}

\newcommand\figref{Fig.~\ref}

\newcommand{\state}{s}
\newcommand{\human}{\mathcal{H}}
\newcommand{\robot}{\mathcal{R}}

\newcommand{\action}{a}
\newcommand{\actionSpace}{\mathcal{A}}

\newcommand{\stateSpace}{\mathcal{S}}
\newcommand{\goal}{g}
\newcommand{\goalinput}{g_\human}
\newcommand{\goalpropose}{g_\robot}
\newcommand{\goalSpace}{\mathcal{G}}
\newcommand{\encoder}{\mathcal{E}}
\newcommand{\intent}{\theta}

\newcommand{\failureset}{\mathcal{F}}
\newcommand{\targetset}{\mathcal{T}}
\newcommand{\targetsetgoal}{\targetset_\goal}
\newcommand{\failuresetgoal}{\failureset_\goal}

\newcommand{\valfuncpolicy}{V_*^\pi}
\newcommand{\valfuncpolicydisc}{V^\pi}

\newcommand{\failure}{{\mathcal{F}}}

\usepackage{fontawesome5}
\newcommand{\policy}{\pi}
\newcommand{\safemargin}{h}
\newcommand{\targetmargin}{l}
\newcommand{\safemargingoal}{h_\goal}
\newcommand{\targetmargingoal}{l_\goal}

\def\BibTeX{{\rm B\kern-.05em{\sc i\kern-.025em b}\kern-.08em
    T\kern-.1667em\lower.7ex\hbox{E}\kern-.125emX}}

\begin{document}
\title{\LARGE \bf Robots that Suggest Safe Alternatives\\
\thanks{\textsuperscript{1}Department of Mechanical and Aerospace Engineering, UC San Diego. Email: \texttt{{hjjeong}@ucsd.edu} \textsuperscript{2}Robotics Institute, Carnegie Mellon University. Email: \texttt{{abajcsy}@cmu.edu}}
}

\author{Hyun Joe Jeong\textsuperscript{1}, Rosy Chen\textsuperscript{2}, and Andrea Bajcsy\textsuperscript{2}
}

\maketitle

\begin{abstract}
Goal-conditioned policies, such as those learned via imitation learning, provide an easy way for humans to influence what tasks robots accomplish. 
However, these robot policies are not guaranteed to execute safely or to succeed when faced with out-of-distribution goal requests.  
In this work, we enable robots to know when they can confidently execute a user's desired goal, and automatically suggest safe alternatives when they cannot. 
Our approach is inspired by control-theoretic safety filtering, wherein a safety filter minimally adjusts a robot's candidate action to be safe. 
Our key idea is to pose alternative suggestion as a safe control problem in \textit{goal} space, rather than in action space. 
Offline, we use reachability analysis to compute a goal-parameterized reach-avoid value network which quantifies the safety and liveness of the robot’s pre-trained policy. 
Online, our robot uses the reach-avoid value network as a safety filter, monitoring the human's given goal and actively suggesting alternatives that are similar but meet the safety specification. 
We demonstrate our Safe ALTernatives (SALT) framework in simulation experiments with indoor navigation and Franka Panda tabletop manipulation, and with both discrete and continuous goal representations. 
We find that SALT is able to learn to predict successful and failed closed-loop executions,  
is a less pessimistic monitor than open-loop uncertainty quantification, 
and proposes alternatives that consistently align with those that people find acceptable. 
\end{abstract}

\section{Introduction}

Imagine that your friend asks you to grab a mug from the top kitchen shelf. Intuitively, you know that trying to reach it will be dangerous because you will drop the mug. Instead of attempting the unsafe task or asking your friend to get it for you, you may naturally suggest an alternative that you can safely accomplish: ``I can't reach your mug, but are you ok with this cup on the lower shelf instead?'' How can we get our robots to operate in same manner? 

In this paper, we want to endow robots with the ability to know when they can confidently execute a user's desired goal and propose safe alternatives when they cannot. Specifically, we study goal-conditioned robot policies \cite{ghosh2019learning} such as those obtained via imitation learning \cite{ding2019goal}. 
While this paradigm has enabled robots to learn complex behaviors and adapt to specified goals online \cite{peng2024preference}, these learned policies can degrade when faced with out-of-distribution goal requests or states \cite{liu2022goalconditionedreinforcementlearningproblems}.
In other words, given a pre-trained goal-conditioned policy, it is hard to ensure that the closed-loop robot behavior will always be safe (e.g., doesn't collide with the environment) and performant (e.g., will successfully pick up the cup) for any new user goal and initial  state. 
Prior works have quantified policy uncertainty \cite{lockwood2022review} or developed out-of-distribution input detectors \cite{yang2024generalized}, but these approaches are only a monitoring mechanism; they don't provide a way for the robot to actively propose alternatives that it can accomplish safely and effectively. 

\begin{figure}
    \centering
    \includegraphics[width=1\linewidth]{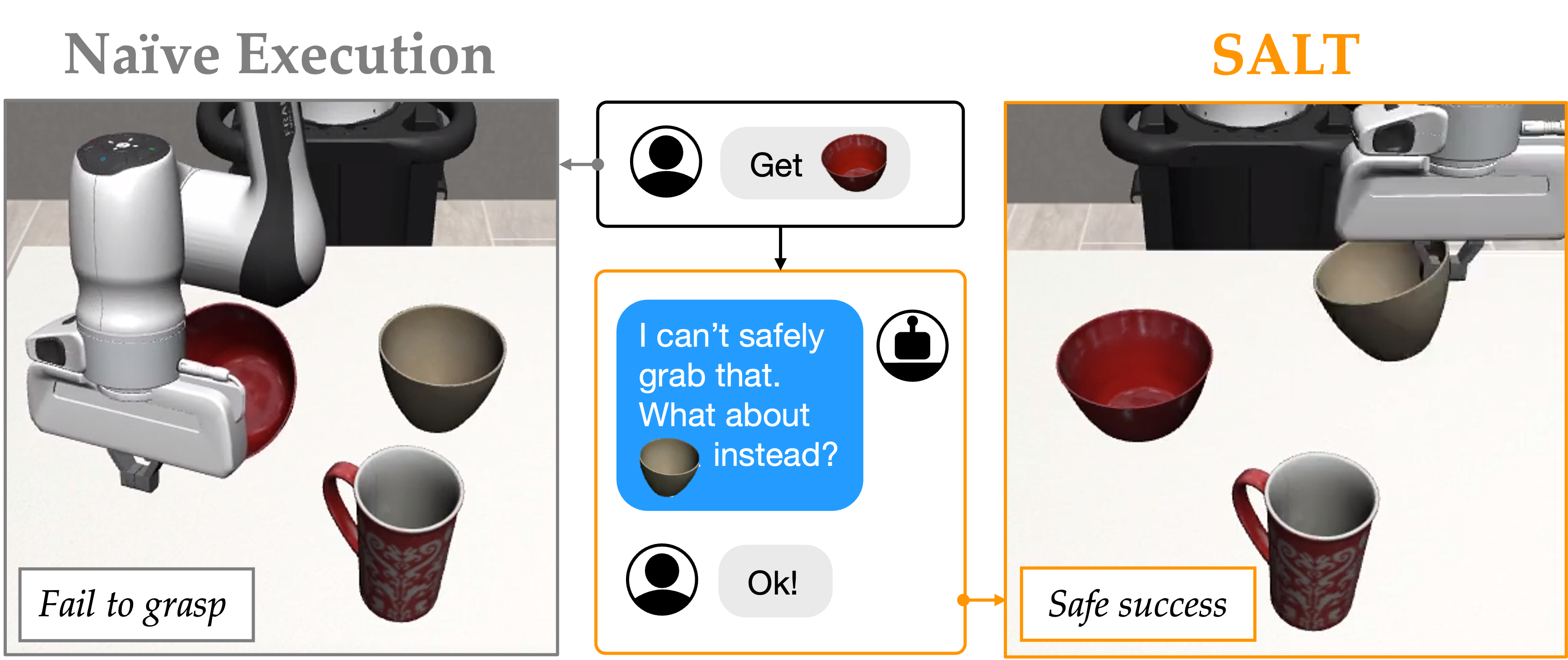}
    \caption{Figure 1: \textbf{Safe ALTernatives (SALT).} If a robot naively executes a user's request, it can fail for a variety of reasons. For example, a request to pick up the red bowl leads the robot to fail to grasp. Our SALT framework enables a robot to detect if it can successfully accomplish a user's original goal; if it cannot, the robot automatically proposes an alternative it can safely succeed at (e.g., get brown bowl). Videos on the project website: \href{https://cmu-intentlab.github.io/salt/}{https://cmu-intentlab.github.io/salt/}.}
    \vspace{-1.2em}
    \label{fig:front-fig}
\end{figure}

To close this gap, we propose that robots suggest safe alternatives. Our approach is inspired by safety filtering techniques from control theory \cite{hsu2023safety}. 
Traditional safety filters take a candidate robot action (e.g., generated by a pre-trained policy) and minimally adjust (i.e., ``filter'') it to be safe. 
The filtering mechanism projects the candidate action onto the safe and live (i.e. goal-reaching) control set, where the control set is computed via methods like Control Barrier \cite{ames2019control} or Control Lyapunov Functions \cite{primbs1999nonlinear}, Hamilton-Jacobi (HJ) reachability analysis \cite{margellos2011hamilton}, or model predictive filters \cite{wabersich2021psf}. However, this action-space filtering does \textit{not} enable the robot to naturally suggest alternatives in a human-verifable way. 
Our key idea is that an
    \begin{quote}
        \centering  
        \textit{alternative suggestion can be modeled as safe control in \textbf{goal space}, rather than action space.}
    \end{quote}
We leverage HJ reachability analysis to synthesize a goal-conditioned reach-avoid value function that is used within our safe control framework. 
This computation is performed once offline and automatically quantifies how capable the robot's pre-trained policy is at accomplishing the task while staying safe, for a suite of possible goal inputs. 
Due to the dimensionality of the problem induced by goal parameterization, we leverage principled but approximate reachability solvers rooted in reinforcement learning \cite{hsu2021safety}, and empirically quantify the error of our learned value function. Online, the reach-avoid value function takes in the current state and a specified goal and determines whether the goal meets the reach-avoid criteria (safe and live). If not, we perform safety filtering over the goals to suggest alternatives.

We call our overall framework for robots that suggest \textbf{S}afe \textbf{ALT}ernatives: \ours.
We demonstrate SALT in simulation experiments grounded in indoor navigation with a vehicle and tabletop manipulation with a Franka Panda arm. We also instantiate our approach with both continuous (i.e., goal locations) and discrete (i.e., objects) goal representations. We find that compared to baselines that only consider the open-loop uncertainty, SALT's understanding of the closed-loop consequences of the robot's policy detect failures 25\% more accurately, and  
the alternative goals that SALT proposes consistently align with those that people find acceptable. 

\section{Related Work}

\para{Safety Filtering}
Safety filters---which detect unsafe actions and minimally modify them---are increasingly popular ways to ensure closed-loop safety \cite{ hsu2024safety, wabersich2023data}. 
The most popular methods are control barrier functions (CBFs) \cite{ames2019control}, Hamilton-Jacobi (HJ) reachability \cite{li2021prediction, driggs2018robust}, and model predictive shielding \cite{brunke2022safe}.  

In this work we build off of HJ reachability due to its ability to handle non-convex target and constraint sets, control constraints and uncertainty in the system dynamics, and its association with a suite of numerical tools including recent neural approximations that scaled safe set synthesis to 15-200 dimensions \cite{fisac2019bridging, bansal2021deepreach}. Our key idea is that by treating the human's goal as a virtual state, we can do safety value function synthesis and safety filtering on the goal (instead of on the actions as is typical). 
This enables the robot to minimally modify the human's desired goal and propose safe alternatives.

\para{Uncertainty Quantification of Learned Robot Policies}
For modular robot policies that utilize an upstream goal or intent estimator, prior works have quantified goal uncertainty \cite{fisac2018probabilistically}, calibrated task plans inferred from language commands \cite{knowno2023} and quantified their execution risk \cite{lidard2024risk}. 
For end-to-end behavior cloned policies, prior works have quantified their generalizability via statistical bounds \cite{vincent2024generalizable}, action uncertainty via temperature scaling \cite{wu2024uncertainty}, and predicted policy success rate via value estimation \cite{gokmen2023asking}. 

Our work uses control-theoretic verification tools to analyze the closed-loop success of a robot's policy. 

\para{Robot Communication of Uncertainty \& Capability} 
Prior works in human-robot interaction have enabled robots to communicate their task uncertainty via dialogue \cite{ren2023robots}, communicate their objectives through motion or haptics \cite{huang2019enabling, mullen2021communicating}, express physical capabilities \cite{kwon2018expressing}, or explain their failures \cite{tagliamonte2024generalizable} to people (see \cite{habibian2023review} for a review). Instead of having robots only explain what they are uncertain about (or ask for help), we enable robots to actively suggest alternatives they can safely accomplish. 

\section{Method: Suggesting Safe Alternatives}

\begin{figure*}[t!]
    \centering
    \medskip
    \includegraphics[width=1
    \linewidth]{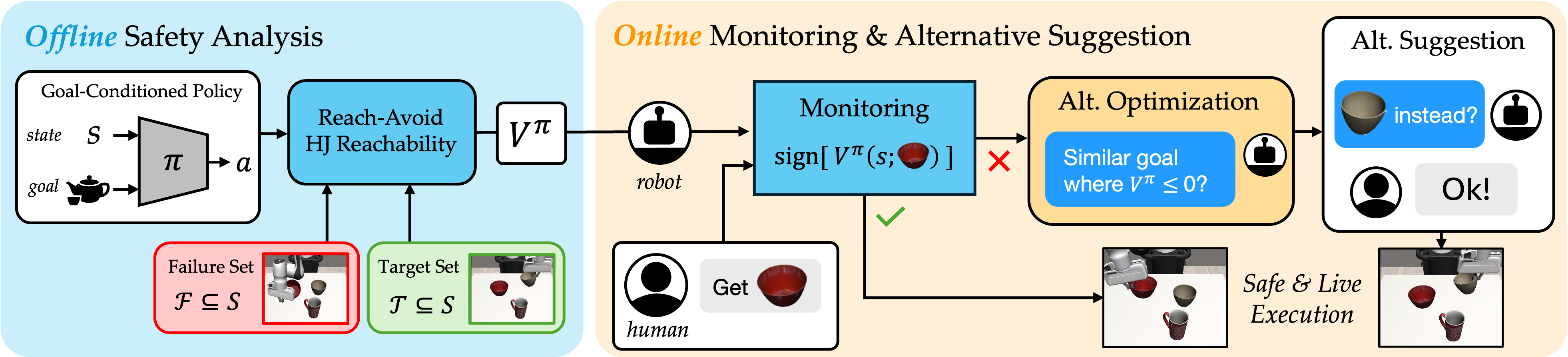}
    \caption{Figure 2: \textbf{Robots that Suggest Safe Alternatives (SALT) Framework.} (Left) Offline, a reach-avoid value network is learned to estimate the safety and liveness properties of a pre-trained goal-conditioned robot policy. (Right) Online, a human inputs a desired goal, which is first monitored by our reach-avoid value function. If the input goal satisfies both safety and liveness, then the policy is executed. Otherwise, the robot solves a safe control problem over alternative goals  (e.g., objects in the scene) to propose an alternative. If the human accepts, then the robot confidently executes on the new goal.}
    \vspace{-1em}
    \label{fig:overview}
\end{figure*}

We want robots to know when they can safely execute a user's given goal and propose safe alternatives when they cannot. 
Our key idea is to formalize alternative suggestion as a safe control problem in goal space.
This goal-space filtering enables the robot to automatically suggest alternatives in a human-verifable way. 
By solving a goal-parameterized reachability problem \textit{offline}, we obtain a reach-avoid value function that the robot uses \textit{online} for filtering.
We call our overall framework for suggesting \textbf{S}afe \textbf{ALT}ernatives,  \ours (summarized in \figref{fig:overview}).

\para{Setup} Let the robot's physical state be  $\state \in \stateSpace$ (e.g., positions, velocities, joint configuration) and the robot's action be $\action \in \actionSpace$. We model the robot's state as evolving via the deterministic discrete-time dynamics $\state_{t+1} = f(\state_t, \action_t)$. The human influences the robot's behavior by specifying a goal $\goal \in \goalSpace$ that can be discrete (e.g., an object a manipulator should pick up) or continuous (e.g., planar position a robot should navigate to). We assume that the robot interprets this goal and executes its behavior based on a pre-trained goal-conditioned policy, $\policy(\state; \goal)$. For example, this could be a behavior cloned policy\footnote{We assume that the policy is deterministic in this work.} trained on a demonstration dataset $(\state, \action, \goal) \sim \mathcal{D}$ consisting of state-action-goal tuples.

\para{\textit{Offline}: Reach-Avoid Analysis of Goal-Conditioned Robot Policies}
We use reachability analysis to automatically verify for which initial states and desired goals can the robot's policy satisfy both safety and liveness constraints. Our core idea is to treat the space of possible goals that the human could ask about at deployment time ($\goal \in \goalSpace$) as a virtual state that has zero dynamics (i.e., $\dot \goal = 0$) during the reachability analysis. This approach, inspired by parameter-conditioned reachability \cite{borquez2023parameter}, 
\textit{enables us to quantify the safety and liveness sensitivity of the robot's policy as a function of all possible goal inputs.} 

For offline analysis, we leverage Hamilton-Jacobi (HJ) reachability. This verification technique is compatible with nonlinear dynamical systems, arbitrary non-convex failure and target set representations, and has a suite of associated numerical tools, from exact grid-based solvers \cite{mitchell2004toolbox} to approximate but scalable techniques such as reinforcement learning \cite{fisac2019bridging, hsu2021safety} and self-supervised learning \cite{bansal2021deepreach}. 

We encode our safety specification via a failure set $\failureset \subset \stateSpace$ (e.g., the object slipped from the gripper) and liveness via a target set $\targetset \subset \stateSpace$ (e.g., the object height must be above the table).
For computation, the target and failure sets are encoded via Lipschitz continuous margin functions $\targetmargin(\cdot)$ and $\safemargin(\cdot)$ respectively: $\targetset := \{ \state \mid \targetmargin(\state) \leq 0 \}$ and $\failureset := \{ \state \mid \safemargin(\state) > 0 \}$.  
One such function is the signed distance to the set boundary. 
Intuitively, these margin functions will measure the ``closest'' the robot's policy $\policy(\state;\goal)$ ever got to violating safety and accomplishing the task given the specified goal $\goal$. We introduce goal-conditioned target and failure sets, $\targetsetgoal$ and $\failuresetgoal$, as well as corresponding 
goal-conditioned margin function, $\targetmargingoal(\state)$ and $\safemargingoal(\state)$. 
With these in hand, we can define the \textit{goal-} and \textit{policy-conditioned} safety value function as:
\begin{align}
    \valfuncpolicy(\state; \goal) = & \min_{\tau \in \{0,1,\hdots\}} \max \Big\{ \targetmargingoal(\xi^{\pi(\cdot;\goal)}_\state(\tau)), \nonumber \\
    & \max_{\kappa \in \{0,\hdots, \tau\}} \safemargingoal(\xi^{\pi(\cdot;\goal)}_\state(\kappa)) \Big\},
\end{align}
where the robot's trajectory starting from state $\state$ and applying policy $\pi(\cdot;\goal)$ is denoted by $\xi^{\pi(\cdot;\goal)}_\state$. 
Intuitively, the outer maximum acts as a mechanism to remember if the robot has ever entered the failure set $\failure$ up to this time (right-hand side) and has satisfied the liveness property (left-hand side). 
If $\xi^{\pi(\cdot;\goal)}_\state$ enters $\failureset$ at any time, then the inner maximum will be positive, and thus the overall value will also be positive. In contrast, if the robot's trajectory never enters $\failureset$, then the overall value will be negative if and only if the robot reaches the target $\targetset$ (which is encoded via the subzero level set of $\targetmargin$). The outer minimum ensures that the value function ``remembers'' these events over the entire time horizon. 
In other words, the value can only be negative if the target $\targetset$ is reached without ever violating the safety constraint $\failureset$ along the way. 

Following prior work \cite{hsu2021safety}, it can be shown that this value function must satisfy the fixed-point \textit{reach-avoid Bellman equation}:
\begin{IEEEeqnarray}{rl}
\label{eq:fixed_point_non_contractive}
    \valfuncpolicy(\state; \goal) = \max\Big\{ \safemargingoal(\state), \ & \min\big\{\targetmargingoal(\state),
    \valfuncpolicy(\state^\pi_+; \goal)\big\}\Big\}.
\end{IEEEeqnarray}
where $\state^\pi_+ := f(\state, \pi(\state; \goal))$ is the next state the robot reaches after applying the control from the goal-conditioned policy $\pi(\cdot; \goal)$ and the $*$ indicates optimality of the reach-avoid value function, $\valfuncpolicy$, computed with a perfect solver.

To scale reach-avoid analysis to higher dimensional state and goal spaces, we leverage the principled time-discounted formulation of Eq.~\ref{eq:fixed_point_non_contractive} introduced in \cite{hsu2021safety}, rendering the reach-avoid problem compatible with reinforcement learning approximations (e.g., Q-learning, REINFORCE \cite{watkins1992q, fisac2019bridging}):
\begin{IEEEeqnarray}{rCl}\label{eq:policy_conditioned_bellman_disc}
    \valfuncpolicydisc(\state;\goal) &=& \gamma\max\Big\{ \safemargingoal(\state), \min\big\{\targetmargingoal(\state),
    \valfuncpolicydisc(\state^\policy_+; \goal)\big\}\Big\} 
    \nonumber \\ 
    && + (1-\gamma)\max\big\{\targetmargingoal(\state),\safemargingoal(\state)\big\}.
\end{IEEEeqnarray}
Here, $\gamma \in [0,1)$ represents the time discount factor, where $\valfuncpolicydisc \rightarrow \valfuncpolicy$ as $\gamma \rightarrow 1$. Note that $\valfuncpolicydisc$ is an over approximation of $\valfuncpolicy$, therefore it will always be more conservative than $\valfuncpolicy$ \cite{hsu2021safety}. 
In the \textbf{SALT} framework, the reach-avoid value function  $\valfuncpolicydisc(\state;\goal)$ is trained offline \textit{before} deploying the robot policy, but will be queried \textit{online} given any new human goal $\goal$ to monitor the robot policy performance and to automatically propose an alternative to the human.    
    
\para{\textit{Online}: Alternative Suggestion as Safe Control in Goal Space} 
Once $\valfuncpolicydisc$ is trained offline, we instantiate the problem of suggesting alternatives \textit{online} as a safe control problem. 
Our key idea is to treat the goal as a virtual action that the robot can minimally modify to ensure the policy will be accomplished safely. 
Specifically, we formalize a ``smooth blending'' safety filter inspired by control barrier functions \cite{ames2019control}, but instead of filtering \textit{actions} as is done typically, we filter \textit{goals}.
Let the human's original goal input be $\goalinput$.
The robot seeks an alternative goal $\goalpropose$ that satisfies:
\begin{IEEEeqnarray}{rCl}
\goalpropose = \arg\min_{\goal\in \goalSpace} ~ d(\encoder(\goal), \encoder(\goalinput); \intent)  \\
\text{ s.t. } \valfuncpolicydisc(\state;\goal) \leq 0, \nonumber 
\label{eq:CBF-goal}
\end{IEEEeqnarray}
where $d(\cdot)$ is a similarity measure, and $\encoder(\cdot)$ is an encoder that maps goal representations to the goal space. The similarity measure is parameterized by human intent $\intent$, since similarity notion differs user by user. We stress that the representation of the goals $\goal$, and the corresponding similarity measure $d(\cdot)$ are a key design decision, and one which we explore in Sec.~\ref{subsec:results-semantic-sim}. 
If the human's original goal $\goalinput$ and robot's initial state $\state$ does \textit{not} satisfy safety and liveness (i.e., $\valfuncpolicydisc(\state;\goalinput) > 0$), then the optimization above will be solved to find an alternative goal $\goalpropose$ that's similar to $\goalinput$. In our experiments, we first start with pose-based representations of $\goal$ and an Euclidean distance function (not conditioned on intent), then explore semantic similarity and goal representations (conditioned on intent). 

\para{Efficiently Optimizing Over Alternatives} If the robot's goal representation is discrete and relatively small, $\goal = \{g_1, g_2, ..., g_N\}$, such as objects on a tabletop, solving the optimization problem in Eq. \eqref{eq:CBF-goal} can be solved exactly via direct enumeration. However, when the goal representation is continuous, Eq. \eqref{eq:CBF-goal} is difficult to solve efficiently because the value function constraint induces a nonlinear optimization problem. 
To efficiently approximate this problem, we take a sampling-based optimization approach. Specifically, we use the objective function to form a Gaussian distribution centered at the human's original goal, biasing sampled alternatives close to the original goal: $\goal \sim \mathcal{N}(\goalinput, \sigma^2), ~ \goal \in \goalSpace$. The mean of this distribution is denoted as $\goalinput$ and it has a small variance, e.g., $\sigma^2 = 0.1$. After generating N samples, we select only those that satisfy $\valfuncpolicydisc(\state,\goal) \leq 0$ and return $\goalpropose$ that is the closest to the original $\goalinput$.   
If none of the N samples satisfy our reach-avoid value function,  
we increase the variance by doubling it and re-sample N new alternative goals. 
We repeat the process of increasing the variance and re-sampling until a suitable goal is found. In the limit, as \( \sigma^2 \rightarrow \infty \), the distribution approaches a uniform distribution over the continuous goal space $\goalSpace$, 
enabling us to consider more and more diverse alternatives. 
If none of these samples meet the requirement, the user is asked to input a new goal. 
In practice, we find that we rarely need to increase the variance and need $N = 100$ samples to find a similar alternative.

\section{Experimental Setup}
\label{sec:setup}


\begin{figure}[t!]
    \centering
    \smallskip
    \includegraphics[width=1\linewidth]{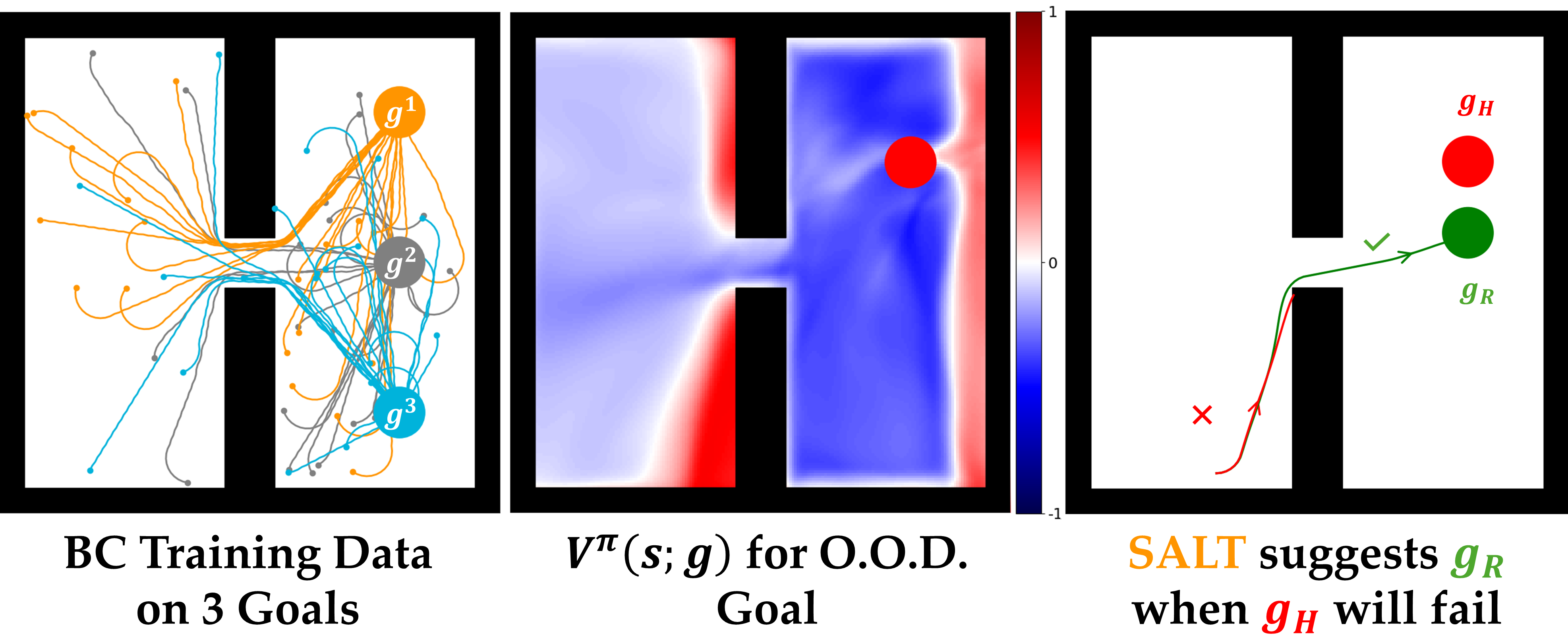}
    \caption{Figure 3: \textbf{SALT Framework in Navigation.} (Left) The BC policy $\policy$ is trained on 50 expert demonstrations for each of the three goals, $\goal^1$, $\goal^2$, and $\goal^3$ (20 per goal depicted). (Center) SALT's reach-avoid value function visualized for all $x,y$, $\phi=0$, and goal $\goal = (3,2)$ that is out-of-distribution (OOD) for $\policy$. (Right) If the person proposes the OOD goal, $\goalinput$, SALT proposes the safe alternative goal, $\goalpropose$. Executing $\policy(\state;\goalinput)$ would lead to an eventual collision (red traj.).}
    \label{fig:navigation}
    \vspace{-1em}
\end{figure}

\para{Environments} We study our SALT framework in two controlled environments: a low-dimensional ground navigation task and a high-dimensional manipulation task.

\parasi{Navigation (4D)} A ground vehicle must navigate to a human-specified goal in an environment with a narrow passage (see Fig.~\ref{fig:navigation}). The robot physical state is the planar position and heading angle $\state := (x, y, \phi)$ and we model the robot as a Dubins' car with fixed linear velocity $v = 1$:
\begin{equation}
\dot{x} = v \sin (\phi), \hspace{2mm} \dot{y} = v \cos (\phi), \hspace{2mm} \dot{\phi} = \omega, 
\end{equation}
where the robot controls angular velocity, $\action := \omega \in [-2,2]$. 
At deployment time, the human can specify a continuous-valued target goal location, $\goal \in \goalSpace \subset \mathbb{R}^2$, anywhere at the end of the corridor ($\goal^x = 3, \goal^y \in [-3,3]$).
To capture this within our base policy and reachability analysis, we augment the state space with a fourth dimension, $(\state, \goal) \in \mathbb{R}^4$ which is the virtual goal state with zero dynamics, $\dot\goal = 0$. 
In turn, this virtual state parameterizes our target set and corresponding margin function: $\targetmargin_\goal(\state) := ||(x,y) - \goal||^2_2 - \epsilon^2$ where $\epsilon = 0.5$. Finally, our safety constraints for reachability analysis are the walls, encoded in $\safemargin_\goal(\state)$ via a signed distance function. 

\parasi{Manipulation (20D)} A tabletop manipulator has to lift a person's desired object: a red mug, a brown bowl, or a red bowl (Fig~\ref{fig:front-fig}). These comprise our discrete goal parameters: $\goal \in \goalSpace := \{\text{RedMug}, \text{BrownBowl}, \text{RedBowl}\}$. We use the robosuite simulation environment \cite{robosuite2020} and the Franka Panda manipulator. 
For both the base policy and for reachability analysis, we model the 20-dimensional robot state $\state \in \mathbb{R}^{20}$ consisting of robot end-effector (EE) pose ($p_{\text{EE}} \in \mathbb{R}^7$, xyz position and quaternion), gripper state (left and right gripper opened or closed $\delta_{\text{L}}, \delta_{\text{R}} \in \{0,1\}$), the pose of the person's desired object ($p_\goal \in \mathbb{R}^7$), the relative position of the EE to the object ($p^{rel}_{\text{EE},\goal} \in \mathbb{R}^3$). We augment the state space with the 1-dimensional desired object id ($\goal \in \goalSpace$).  
The robot's action space $\action \in \mathbb{R}^7$ controls the EE linear (x,y,z) and angular (roll, pitch, yaw) velocity and gripper open and close. 
All control inputs are bounded to a magnitude of $[-1, 1]$. The target margin function is $\targetmargin_\goal(\state) = ||p^z_\goal - 0.8||_2^2 - \epsilon$, encoding the z-distance of the object from the target height above the table (0.8 m), and the failure margin function is $\safemargin_\goal(\state) = \min\{\min(\delta_{\text{L}}, \delta_{\text{R}}) < 0.001,||p^{rel}_{\text{EE},\goal}|| - 0.1\}$, which measures if the robot is gripping without the object in hand.

\para{Base Robot Policy ($\policy$)} For all environments, the robot's base policy is trained via behavior cloning (BC).
For \textbf{navigation}, expert demonstrations $(\state, \action, \goal) \sim {\mathcal{D}}$ are collected by using an optimal control law obtained from a reach-avoid computation obtained via the optimizedDP solver \cite{https://doi.org/10.48550/arxiv.2204.05520}. 
We pre-compute three separate optimal controllers for three goal locations, $\goal_{\mathcal{D}} \in \{(3, -3), (3,0), (3, 3)\}$ and obtain 50 successful expert trajectories per goal. The BC policy is a 3-layer MLP with 128 neurons per layer and RELU nonlinearity and trained via the mean squared error loss $\mathcal{L}_{BC}(\mathcal{D}) = \mathbb{E}_{(s^j, a^j, g^j) \sim \mathcal{D}} \left\| \pi(s^j; g^j) - a^j \right\|_2^2$, where $j$ is each state-goal-action pair. We train for 500 epochs with a 100 epoch early stop patience. 
In \textbf{manipulation}, we collected demonstrations in the robomimic \cite{robomimic2021} environment via expert human teleoperation, and obtained 100 successful demonstrations per object. 
The BC policy is a 2-layer MLP with 1024 neurons per layer and ReLU nonlinearity. It is trained with the AdamW \cite{loshchilov2019decoupledweightdecayregularization} optimizer and for 100 epochs. Every 5 epochs, we rollout the policy 50 times, and use the policy with the highest success rate.


\para{Reach-Avoid Value Learning ($\valfuncpolicydisc$)} We use the off-the-shelf reach-avoid reinforcement learning (RARL) solver from \cite{hsu2021safety} to approximate the value function in Eq.~\eqref{eq:policy_conditioned_bellman_disc}. 
We train on a single NVIDIA 4090ti GPU for 400k epochs total, checkpointing at every 50k epochs. We anneal the discount factor $\gamma$ from 0.9 to 0.9999 throughout training, use a 3-layer MLP with 512 neurons per layer and tanh nonlinearity, and the AdamW optimizer.
We warm-up the value network for 50k iterations by sampling random points in the state space to properly learn $\targetmargin_\goal(\state)$ and $\safemargin_\goal(\state)$. 400 and 500 max episode steps are set for navigation and manipulation, respectively. Note that the each episode step is 0.1 seconds, so the time horizons for navigation and manipulation are 40 seconds and 50 seconds respectively. In the navigation example we use a first-principles Dubins' dynamics model to obtain trajectory rollouts, while in the manipulator setting we use the MuJoCo \cite{mujoco2012} simulation environment.


\section{Experimental Results}
\label{sec:results}

To understand each component of \ours, we study four questions in simulation experiments: (1) how accurate is the reach-avoid value function approximation?, (2) how does \ours compare to alternative runtime monitoring schemes (e.g., ensembles)?, (3) how should \ours measure ``similar'' alternative goals?, and (4) how aligned are the alternatives that \ours proposes with human-acceptable ones?

\subsection{How Accurate is SALT's Reach-Avoid Value Function?}

\para{Metrics} To evaluate the reliability of the learned value network, we measure the true success rate (\textbf{TSR}: network predicts the policy can safely accomplish the task and in reality it can), true failure rate (\textbf{TFR}: network predicts the policy \textit{cannot} safely accomplish the task and in reality it cannot), false success rate (\textbf{FSR}: network predicts the policy \textit{can} safely accomplish the task but in reality it cannot), and false failure rate (\textbf{FFR}: network predicts the policy \textit{cannot} safely accomplish the task but in reality in can). $F_1$-Score represents the predictive performance of a binary classifier (1.0 indicates perfect precision).

\para{Evaluation Approach} For the low-dimensional (4D) navigation example, we evaluate our approximation for a dense gridding of the state space (50x50x20x20). 
For the high-dimensional manipulation example, exhaustive gridding is not feasible. 
Thus, we randomly sample initial physical states $\state$ and goals $\goal$, perform a rollout in our simulator, and check if the value of the network accurately reflects policy execution outcome (safe success, or not). We do this on 1,000 initial $(\state, \goal)$ pairs sampled near the zero level set of the approximate value network ($\valfuncpolicydisc(\state,\goal) \approx 0$), since a accurate boundary matters for monitoring, and 
across 10 random seeds. 

\para{Results} Table \ref{tab:valfunc_eval} shows accuracy metrics for navigation and manipulation. 
In the navigation setting, we obtain a high-quality value function with about $95\%$ average true success and true failure rate. 
We see a $75\%$ TSR + TFR value function approximation accuracy in the manipulation task and hypothesize this is due to the much higher dimensional system and complexity of the grasping and pickup task. 
Future work should investigate post-hoc adjustment techniques to further minimize the FFR and FSR rates (e.g., \cite{lin2023verification}). 

\begin{table}[t!]
    \centering
    \medskip
    \resizebox{\columnwidth}{!}{%
    \begin{tabular}{l|c c c c|c}
        \toprule
        Environment & \multicolumn{1}{c}{TSR (\%)} & \multicolumn{1}{c}{TFR (\%)} & \multicolumn{1}{c}{FSR (\%)} & \multicolumn{1}{c|}{FFR (\%)} & \multicolumn{1}{c}{$F_1$-Score} \\
        \midrule
        \textbf{Navigation} & 55.51 & 39.48 & 1.28 & 3.72 & 0.96 \\  
        \textbf{Manipulation} & 59.98 ($\pm$2.10) & 14.55 ($\pm$1.43) & 12.64 ($\pm$1.28) & 12.83 ($\pm$1.13) & 0.82 \\  
        \bottomrule
    \end{tabular}
    }
    \vspace{0.1em}
    \caption{TABLE I: \textbf{Quality of the SALT's value function approximation.} Confusion matrix for navigation and manipulation. Navigation was rolled out exhaustively on a grid, and manipulation sampled 1,000 initial conditions near the zero level boundary of the value network across 10 randomized seeds.}
    \label{tab:valfunc_eval}
    \vspace{-1em}
\end{table}

\subsection{What is the Benefit of SALT as a Runtime Monitor?}

\begin{table*}[ht!]
    \centering
    \medskip
    \resizebox{\textwidth}{!}{%
    \begin{tabular}{l|c c c c|c|c c c c|c}
        \toprule
        & \multicolumn{5}{c|}{\textbf{Navigation}} & \multicolumn{5}{c}{\textbf{Manipulation}} \\
        Method & \multicolumn{1}{c}{TNR \% ($\uparrow$)} & \multicolumn{1}{c}{TPR \% ($\uparrow$)} & \multicolumn{1}{c}{FPR \% ($\downarrow$)} & \multicolumn{1}{c|}{FNR \% ($\downarrow$)} & \multicolumn{1}{c|}{$F_1$-Score ($\uparrow$)} & \multicolumn{1}{c}{TNR \% ($\uparrow$)} & \multicolumn{1}{c}{TPR \% ($\uparrow$)} & \multicolumn{1}{c}{FPR \% ($\downarrow$)} & \multicolumn{1}{c|}{FNR \% ($\downarrow$)} & \multicolumn{1}{c}{$F_1$-Score ($\uparrow$)} \\
        \midrule
        \ensemble & 65.97 ($\pm$0.88) & 1.92 ($\pm$0.38) & 12.31 ($\pm$0.48) & 19.80 ($\pm$0.52) & 0.80 & 34.61 ($\pm$1.68) & 27.45 ($\pm$1.96) & 32.65 ($\pm$1.27) & 5.26 ($\pm$0.89) & 0.65 \\
        \rl & 53.35 ($\pm$1.34) & 5.95 ($\pm$0.74) & 4.87 ($\pm$0.60) & 35.78 ($\pm$1.26) & 0.72 & 64.47 ($\pm$1.61) & 4.67 ($\pm$0.68) & 8.85 ($\pm$1.01) & 22.01 ($\pm$1.21) & 0.81 \\
        \ours (ours) & 57.00 ($\pm$1.97) & 37.34 ($\pm$2.03) & 1.28 ($\pm$0.15) & 4.38 ($\pm$0.78) & \textbf{0.95} & 61.21 ($\pm$1.91) & 13.23 ($\pm$1.56) & 13.23 ($\pm$1.23) & 12.33 ($\pm$0.85) & \textbf{0.83} \\
        \bottomrule
    \end{tabular}
    }
    \vspace{0.1em}
    \caption{TABLE II: \textbf{Evaluating Runtime Monitors.} Empirical safety metrics for navigation and manipulation tasks. The evaluation was done on 1000 random initial states across 10 randomly seeded iterations.}
    \label{tab:empirical_safety}
    \vspace{-1em}
\end{table*}

\para{Baselines} We compare our reachability-based monitor to two baselines: \ensemble and \rl safety monitors. 
Following \cite{lakshminarayanan2017simple}, we use an \ensemble of behavior cloned policies as an \textit{open loop} monitor: high ensemble disagreement is a measure of uncertainty in the robot's action prediction. If the disagreement exceeds a threshold, then the robot stops and asks for help. 
For both environments, we use $M = 5$ policies as ensemble members, and take the variance $\sigma^2$ of the action prediction as the uncertainty measure; the robot stops when  $\sigma^2 > \epsilon$. We use $\epsilon = 2.5$ for navigation and $\epsilon = 0.0175$ for manipulation, which are heuristically tuned for lowest FSR and FFR. 
Similar to our approach, \rl is a \textit{closed loop} safety monitor whose value function captures long-term outcomes of executing the base policy. 
However, the two methods differ in their optimization objective: \rl computes the value via the typical \textit{expected sum} of discounted rewards used in reinforcement learning, while \ours uses the reach-avoid objective which remembers the \textit{closest} the robot ever gets to safety and liveness violations (as in Eq.~\eqref{eq:policy_conditioned_bellman_disc}). 

\para{Metrics} We measure the accuracy of each monitor stopping to alert the human. 
We once again randomly sample $(\state, \goal)$ pairs,  
roll our the policy to obtain the ground-truth success or failure label, and then compare each monitor to this label.  
We define a 2x2 confusion matrix of monitor predictions (flag raised or not) and actual robot outcomes (success or fail). 
True negative rate (TNR) is when the monitor does not raise a flag and the robot executes successfully, true positive rate (TPR) is when the monitor raises a flag and the robot would have actually failed, false positive rate (FPR) is when the monitor raises a flag unnecessarily, and false negative rate (FNR) is when when the monitor did not raise a flag when it should have (robot failed). 

\begin{figure}[t!]
    \centering
    \medskip
    \includegraphics[width=1\linewidth]{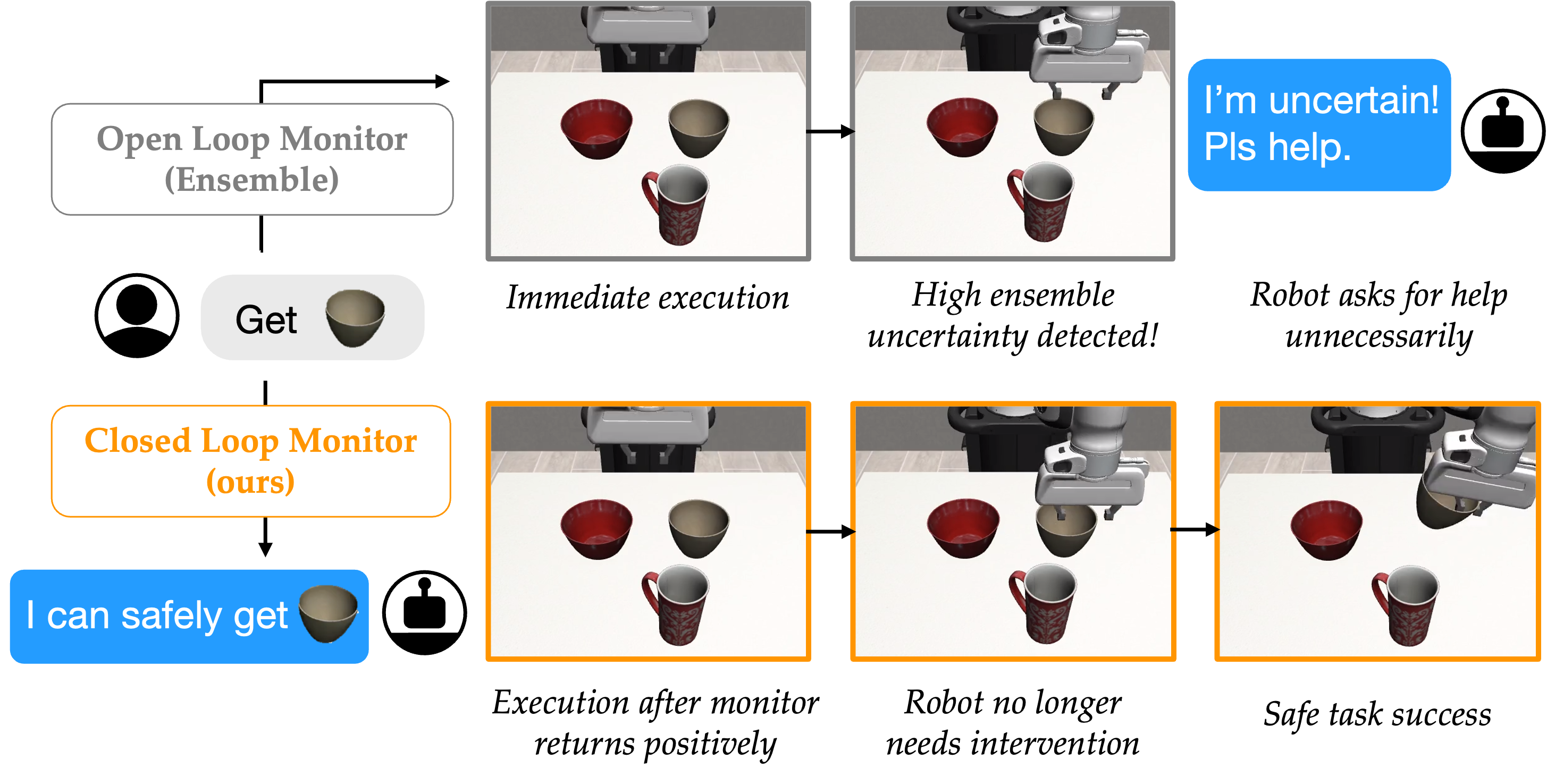}
    \caption{Figure 4: \textbf{Open Loop vs. Closed Loop Monitoring.} \ensemble detects uncertainty before grasping and asks for help unnecessarily. \ours's closed loop monitoring first checks the safety and liveness via the $\valfuncpolicydisc$, then executes confidently.}
    \label{fig:loops}
    \vspace{-1em}
\end{figure}

\para{Results}
Results are in Table \ref{tab:empirical_safety}. In the \textbf{navigation} setting, the open loop \ensemble monitor had the lowest TPR (1.92\%) while having the highest FPR (12.31\%). 
This implies that open loop monitors are overly \textit{pessimistic}, asking for help unncessarily.
Although \rl has a higher TPR compared to \ensemble, it is overly \textit{optimistic} compared to SALT's reach-avoid value function, as shown from the relatively high FNR rate (35.78\%). 
\ours has the highest TPR (37.34\%) and lowest FNR (4.38\%), indicating that it can accurately predictor prior to the robot's execution if the behavior will succeed safely or fail. A visual of SALT's $\valfuncpolicydisc$ is shown in center of Figure~\ref{fig:navigation}.

In  \textbf{manipulation}, \ensemble had the highest TPR (27.45\%) but also the highest FPR (32.65\%). 
In total, \ensemble triggered human help around 60\% of the time, when it should have been triggered less than half of the time.
Figure \ref{fig:loops} shows an example of this pessimism: \ensemble asks for help mid-execution even though the robot was capable of doing the task safely. 
Since actions are being evaluated at every timestep, \ensemble can only know that it is uncertain, rather than describe a high-level alternative that would make it confident.
In contrast, ours checks pre-execution if the closed-loop behavior is predicted to be safe and live, and then no longer requires human supervision during execution. 
Finally, comparing closed loop monitors, \ours is consistently a more reliable safety monitor compared to \rl, having a high TPR+TNR of 74.44\% (vs. 69.14\%) and a lower FNR of 12.33\% (vs. 22.01\%).

\subsection{How Should SALT Reason About ``Similar'' Alternatives?}
\label{subsec:results-semantic-sim}

We hypothesize that suggesting similar alternatives may require a \textit{semantic} representation of the goals---capturing visual or functional properties of the object---and a corresponding similarity measure.   
In this section, we ablate the encoder models $\encoder(\cdot)$ and similarity measure $d(\cdot)$ used within our \ours framework.
We also study how similar alternatives change as a function of human intent $\theta$ in Equation~\ref{eq:CBF-goal}. 

\para{Methods: Encoders, Intents, and Similarity Measures} 
We investigate three methods that leverage different goal representations and similarity measures: \BERT, \GPT, and \SIRL. 
In our experiments, \BERT uses a textual description of the goal (e.g., as returned by a semantic object detector) and measures similarity via the cosine similarity between the textual embedding of the human's original goal ($\goalinput)$ and any alternative goal ($\goal$). 
We use the pre-trained BERT \cite{devlin2019bertpretrainingdeepbidirectional} sentence-transformer model to obtain the textual embedding. 
Mathematically, given a language description $\mathcal{L}_\goal$ of a goal $\goal$ and a textual description of the intent $\mathcal{L}_\intent$, our encoder produces an embedding vector $\encoder_{\text{BERT}}(\mathcal{L}_\goal; \mathcal{L}_\intent) = \vec{w}_\goal$ and our similarity measure is $d := \vec{w}_\goal \cdot \vec{w}_{\goalinput} / ||\vec{w}_\goal||~||\vec{w}_{\goalinput}|| $. 
\GPT Fuzzy Matching \cite{duan2024aha} also uses a textual representation of the goal and intent, but uses a pre-trained large language model (LLM) to directly reason about the semantic similarity (without looking at the embedding similarity). In our experiments, we use GPT-4o \cite{openai2024gpt4ocard} as our language model. Mathematically, $d := \GPT(\mathcal{L}_\goal, \mathcal{L}_{\goalinput}; \mathcal{L}_\intent, \mathcal{P})$, where the prompt to the LLM is $\mathcal{P}=$ ``The user intends to $\mathcal{L}_\theta$. Given $\mathcal{L}_{\goalinput}$, which item is the closest related to it?''.

Unlike \GPT and \BERT (which use pre-trained language models), \SIRL requires training a \textit{personalized} representation which explicitly learns an end-user's notion of similarity from their preference data, enabling us to study how a personalized model of similarity influences our \ours framework compared to pre-trained models that are not fine-tuned on individual data. This model uses a privileged, hand-engineered feature space $\Phi_\goal \in \mathbb{R}^m$ as input based on any given goal object $\goal$; for example, in our experiments, given a $\goal$ is a red cup, $\Phi_\goal$ would be a 20-dimensional vector consisting of the object's RGB color values and functional and material properties. 
We train an intent-parameterized encoder $\encoder_{\SIRL}(\Phi(\goal); \intent) = \phi^\intent_\goal$ via contrastive learning, which returns an embedding vector $\phi^\intent_\goal \in \mathbb{R}^n, ~n < m$ that represents the most relevant features of a goal given the user's intent.
Finally, \SIRL measures similarity via L2 distance in embedding space: $d := ||\phi^\intent_\goal - \phi^\intent_{\goalinput}||_2$.



\begin{figure*}[t!]
    \centering
    \includegraphics[width=0.8\linewidth]{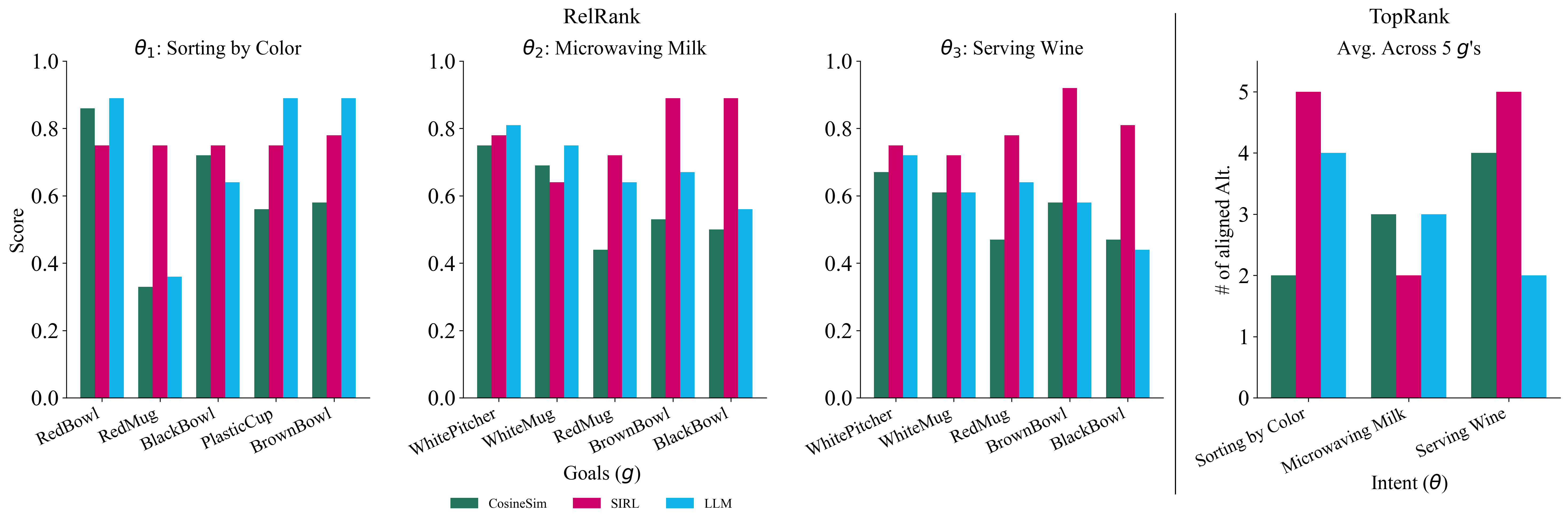}
    \caption{Figure 5: \textbf{Accuracy of Different Similarity Measures.} (Left Three Plots) \texttt{RelRank} scores for 5 aligned $\goalinput$ are scored across baselines and intents. (Right Plot) \texttt{TopRank} scores for 5 aligned $\goalinput$ are scored across baselines and intents.}
    \label{fig:reldist}
    \vspace{-1em}
\end{figure*}

\para{Implementation Details} \SIRL learns relevant similarity features by asking the end user to select the two most similar goals given a triplet of goals. Throughout this section, we use simulated human data for training SIRL and evaluation, enabling us to have access to a ground-truth representation of the human's notion of similarity given the intent. 
\SIRL is trained on triplets $\mathcal{D}_\intent = \{(\Phi^i_{\goal_1}, \Phi^i_{\goal_2}, \Phi^i_{\goal_3})\}^K_{i=1}$, which are feature spaces corresponding to three distinct goals in the environment given a human intent $\intent$. The simulated human ranks the two most similar ones using their ground-truth intent-relevant features. 
We train the encoder $\encoder_{\SIRL}(\Phi(\goal); \intent) = \phi^\intent_\goal$ to minimize the loss from \cite{bobu2023}: 
\begin{equation}
    \mathcal{L}(\phi^\intent_\goal) = \sum_{i=1}^{|\mathcal{D}_\intent|} \mathcal{L}_{trip}(\Phi_{g}^i, \Phi_{g_+}^i, \Phi^i_{g_-}) + \mathcal{L}_{trip}(\Phi_{g_+}^i, \Phi_{g}^i, \Phi^i_{g_-}) \nonumber 
\end{equation}
where $\Phi_{g}^i$ and $\Phi_{g_+}^i$ are ranked as most similar and $\Phi_{g_-}^i$ is most dissimilar. $\mathcal{L}_{trip}(\Phi_{g}^i, \Phi_{g_+}^i, \Phi^i_{g_-})$ is the triplet loss \cite{balntas2016learning} that uses $\Phi_{g}^i$ as the anchor, $\Phi_{g_+}^i$ as the similar example and $\Phi^i_{g_-}$ as the dissimilar example. 
This loss function pushes together embeddings for similar objects with respect to the intent-relevant features while pushing apart embeddings for dissimilar objects.

\para{Evaluation Setup} We use 10 kitchen objects (e.g., cups, bowls, mugs, pitchers, teapots, ramekins) from the Google Research dataset \cite{downs2022googlescannedobjectshighquality} and study  three increasingly complicated user intents: $\theta_1 =$ \textit{Sorting Kitchen by the Same Color}, $\theta_2=$ \textit{Microwaving Soup}, and $\theta_3=$ \textit{Serving Wine at Formal Dinner}. For each intent, we select 5 initial goal objects from the dataset that make sense for the intent. 
Given one of these five objects, we query each method for its similarity score compared to all other 9 objects. We determine the ground-truth similarity score using a simulated human with privileged knowledge about their intent-relevant features.

\para{Metrics} We want to measure how well the distance functions in each method captures the relative distance between any pairs of objects. We use two metrics: \texttt{TopRank} returns one if the most similar item outputted by a similarity measure is the most similar item for the simulated human (and zero otherwise); \texttt{RelRank} returns a real-valued score (between 0 to 1) which measures how correctly a method ranks all other goals relative to a given goal\footnote{Our metric is inspired by from Kendall's rank correlation coefficient \cite{stepanov2015kendallcorrelationcoefficient}. 
However, while Kendall’s penalizes (with -1) for every incorrectly ordered pair, our metric does not (assigns 0). This modification is appropriate for our context, where the top rankings are of more practical importance (since we want to return \textit{maximally} similar alternatives).  Thus, we should not punish the method for having noisy bottom rankings since the differences may not represent meaningful distinctions.} 
. Mathematically, let $\goalinput$ be a given goal and let $\mathcal{G}^*_{\goalinput} = \{g_1, g_2, ... g_n\}$ be a list of all other $n$ goals ranked by how similar they are to $\goalinput$ by the simulated human. 
We define  $\mathcal{G}^\omega_{\goalinput}$ to be a list of all the $n$ goals ranked by their similarity to $\goalinput$ by any method, $\omega \in \{ \SIRL, \BERT, \GPT \}$. 
Let $r_{\mathcal{G}^*_{\goalinput}}(g_i)$ be the rank position of goal $g_i$ in the ordered list $\mathcal{G}^*_{\goalinput}$ and  let $R_{\mathcal{G}^*_{\goalinput}}(g_i,g_j) = \text{sign}[r_{\mathcal{G}^*_{\goalinput}}(g_j) - r_{\mathcal{G}^*_{\goalinput}}(g_i)]$. 
Here, $R_{\mathcal{G}^*_{\goalinput}}(g_i,g_j)$ 
returns 1 if $g_i$ is more similar to $\goalinput$ than $g_j$ and -1 for the converse (our lists assume no ties). 
Finally, we define an indicator function 
$\mathsf{1}[R_{\mathcal{G}^*_{\goalinput}}(g_i,g_j) = R_{\mathcal{G}^\omega_{\goalinput}}(g_i,g_j) ]$
that returns 1 if the two lists have the same relative rankings of the goals 
and 0 otherwise. 
The \texttt{RelRank} metric is then:
\begin{align}
\texttt{RelRank}&(\mathcal{G}^*_{\goalinput}, \mathcal{G}^\omega_{\goalinput}) = \\ \nonumber
\frac{1}{\binom{n}{2}} &\sum_{1 \leq i < j \leq n} 
\mathsf{1}[R_{\mathcal{G}^*_{\goalinput}}(g_i,g_j) = R_{\mathcal{G}^\omega_{\goalinput}}(g_i,g_j) ]. \nonumber
\end{align}

\para{Results: Most Similar Goal Accuracy} 
In the right of Figure~\ref{fig:reldist} we show the \texttt{TopRank} results averaged across all five goals per each intent. \SIRL most consistently ranks the human's preferred goal as most similar (in total, 12) compared to \BERT and \GPT (9 for both). Furthermore, the explicit training of \SIRL makes it more robust to intents while \GPT and \BERT (parameterized by the intent) sometimes ignore critical intent features. (e.g., for $\theta_1=$\textit{Sorting Kitchen by Color} and $\goalinput=\text{RedMug}$, both methods return White Mug, neglecting color). We hypothesize that this occurs because the models are only approximately optimal in their rankings and selecting the top-1 choice requires precision; for example, 
we observed that the ground-truth top choice typically appeared within the top-3 results of each method. 

\para{Results: Overall Similarity Measure Accuracy} We report the \texttt{RelRank} metric (which quantifies the overall performance of a similarity measure) in the three plots left of Figure \ref{fig:reldist} for each intent $\theta_1,\theta_2, \theta_3$.  
Across all intents and goals, \SIRL's similarity measure is more consistently accurate compared to \BERT and \GPT, and is the best performing measure for intent $\intent_3$. 
This is because \SIRL is optimized to solve a personalized metric learning problem---identifying an embedding space that understands similarity according to the user's internal state---while the other two approaches that use pre-trained models have an implicit semantic understanding of similarity. As the intents become increasingly complicated (e.g., $\intent_3=$\textit{Serving Wine}, \BERT and \GPT 's pre-trained similarity struggle to capture how the human evaluates these alternatives, while \SIRL maintains performance due to its privileged access to human's internal states.  Between the two pre-trained approaches, \GPT typically out-performs \BERT in terms of accuracy.


Our main takeaway is that personalized preference data enables more accurate goal representations and better similarity measures, particularly for complex intents. 
However, off-the-shelf language models and LLM Fuzzy Matching can still be valuable semantic similarity measures that require no additional training data and provide a more naturalistic and intuitive interface for people to specify their goals (e.g., via language rather than explicit featurizations of the world). 

\subsection{How Acceptable are SALT's Proposed Alternatives?}

Finally, we study \ours's overall performance at detecting when the robot can safely accomplish a desired goal and suggesting a similar safe alternatives when it cannot. 

\para{Alternatives Dataset} To measure how acceptable are \ours's alternatives, we obtained a validation dataset of initial goal ($\goalinput$) and acceptable alternative pairs annotated by 20 expert users from labs at Carnegie Mellon and UC San Diego. In the manipulation setting, people were shown a tabletop with three objects (as in \figref{fig:front-fig}) with two distinctive intents: $\theta_1=$ \textit{Drinking Soup} and $\theta_2=$ \textit{Sorting Kitchen by the Same Color}. These users were asked which alternative objects would be acceptable given an initial object.

\para{Approach} For each initial goal $\goalinput$ in the validation dataset, we queried \ours to obtain our algorithm's suggested alternatives. We perform this across a suite of initial robot states $\state$. We queried 1,000 random initial conditions for each goal and saved our algorithm's suggestion, $\goalpropose$. Note that \ours returns the initial goal (i.e., $\goalpropose = \goalinput$) if it is safe and live.

\para{Metrics} We measure alternative alignment (\%): given an initial goal $\goalinput$, if the alternative goal proposed by \ours matches or if the input goal is safe and live, then alignment is a success. We compute mean and standard deviation across all users and initial $\state$, and report results per initial goal. We compare alignment across several distance measures: \Euclid, \BERT, \SIRL, and \GPT.

\begin{figure}[t!]
    \centering
    \includegraphics[width=0.9\linewidth]{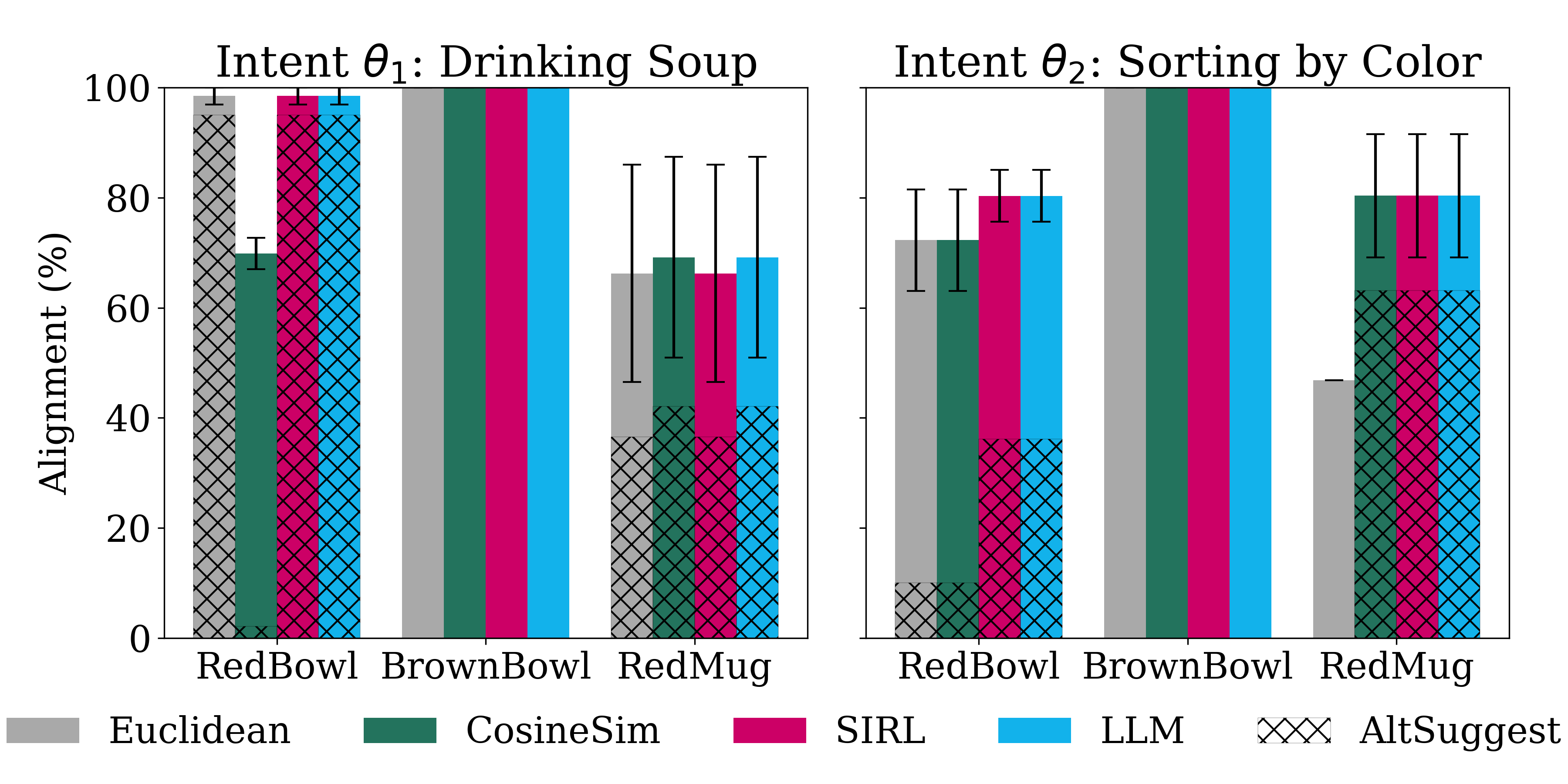}
    \caption{Figure 6: \textbf{User Alignment Success Across 1,000 Initial Conditions.} \Euclid is in grey, \BERT in green, \SIRL in pink, and \GPT in blue. Scenarios where the initial  goal $\goalinput$ was \textit{not} safe and thus the robot had to suggest an alternative are denoted as AltSuggest. \SIRL and \GPT consistently out-perform \Euclid and \BERT when it comes to guiding \ours to generate safe and aligned alternatives.}
    \label{fig:alignment}
    \vspace{-1em}
\end{figure}

\para{Results: Quantitative} 
Alternative alignment results are shown in Figure~\ref{fig:alignment}. 
On aggregate, we find that \ours can detect if the original goal was safe and suggest safe alternatives that strongly align with human preferences (alignment scores between 70\%-100\% when using the \SIRL or \GPT objective functions.   
We note that when $\intent_1=$\textit{Drinking Soup}and $\goalinput=\text{RedMug}$, 35\% of the users wanted neither alternative, dropping the maximum possible alignment rate. 

Furthermore, we break down our results into only those scenarios where the human's initial goal ($\goalinput$) was \textit{not} safe, and thus the robot \textit{had} to suggest an alternative. 
The alignment scores for these scenarios are shown in Figure~\ref{fig:alignment} in a cross-hatched pattern (called AltSuggest). Note that for $\goalinput=\text{BrownBowl}$, the robot is always capable of grabbing this initial object safely, and thus it is not part of the AltSuggest breakdown.
First, we see that \Euclid has low alignment for most of the AltSuggest scenarios, highlighting the need for semantic similarity measures. 
For intent $\intent_1=$\textit{Drinking Soup} and $\intent_2=$\textit{Sorting by Color}, \BERT's alignment rate drops significantly (to 2.13\% and 10\% respectively) when $\goalinput=\text{RedBowl}$, indicating that this similarity measure is not capable of making the closed-loop system suggest safe \textit{and} aligned alternatives. 
Consistent with Section~\ref{subsec:results-semantic-sim}, we see that when \ours uses \SIRL and \GPT, the overall system is able to consistently return safe and similar alternatives that align with an end-user's notion of similarity. 

\para{Results: Qualitative} 
Figure \ref{fig:front-fig} shows our algorithm in manipulation where a user first asks for the red bowl to be picked up with the intent $\theta_1=$ \textit{Drinking Soup}. Realizing the it would likely mis-grasp the red bowl and fail, SALT proposes to pick up the brown bowl with \SIRL (right, Figure \ref{fig:front-fig}), and safely completes the task.

\section{Conclusion}
In this work, we propose SALT: a framework for robots to monitor if their goal-conditioned policies can safely accomplish a given goal, and automatically suggest safe alternatives when they cannot. By actively proposing a safe alternative pre-execution, we can not only minimize possible safety violations, but also human intervention efforts. For safety, we find that open loop monitors that interpret safety as uncertainty cannot differentiate clearly between a safe and unsafe state, while our closed-loop monitor reliably predicts success and failure. 
One limitation of our work is that we rely on privileged state information about the environment as well as goal representations to quantify safety and suggest alternatives. 
Another limitation is the use of a simulated human during our similarity measure evaluation, since it may not accurately model real user preferences. 
We are excited about further verifying this with a rigorous user study. 
In future work, we seek to investigate image-based goal-conditioned policies, as well as image representations of goals and similarity queries. 
One exciting future work is reasoning about task-level alternatives, rather than high-level alternatives like goals.

\section*{Acknowledgment}
This work is supported by NSF Award \#2246447 and the Robotics Institute Summer Scholars (RISS) program at Carnegie Mellon University. The authors would like to thank Ken Nakamura on his insights on reach-avoid RL, Ravi Pandya on his help with robosuite and robomimic, and Lasse Peters for his discussions at various stages in the project.

\bibliographystyle{IEEEtran}
\bibliography{Bib/bajcsy, Bib/reachability, Bib/misc, Bib/fisac}

\end{document}